\documentclass[10pt,twocolumn,letterpaper]{article}

\usepackage[pagenumbers]{cvpr}
\usepackage{graphicx}
\usepackage{amsmath}
\usepackage{amssymb}
\usepackage{booktabs}
\usepackage{multirow}
\usepackage{array}
\usepackage{xcolor}
\usepackage{algorithm}
\usepackage{algorithmic}
\usepackage{gensymb}
\usepackage[breaklinks=true,bookmarks=false]{hyperref}

\begin{document}

\title{Gait Recognition via Deep Residual Networks and Multi-Branch Feature Fusion}

\author{Yabo Luo\textsuperscript{1} \quad Xiaoyun Wang
\textsuperscript{1} \quad Cunrong Li\textsuperscript{1}\\
\textsuperscript{1}School of Mechanical and Electrical Engineering, Osh State University
}

\maketitle

\begin{abstract}
Gait recognition has emerged as a compelling biometric modality for surveillance and security applications, offering inherent advantages such as non-intrusiveness, resistance to disguise, and long-range identification capability. However, prevailing approaches struggle to comprehensively capture and exploit the rich biometric cues embedded in human locomotion, particularly under covariate interference including viewpoint variation, clothing change, and carrying conditions. In this paper, we present a high-precision gait recognition framework that deeply extracts and synergistically fuses gait dynamics with body shape characteristics through a multi-branch architecture grounded in deep residual learning. Specifically, we first employ the High-Resolution Network (HRNet) to perform robust skeletal keypoint estimation, preserving fine-grained spatial information even under low-resolution inputs. We then construct three complementary feature branches---body proportion, gait velocity, and skeletal motion---from the extracted pose sequences. A 50-layer Residual Network (ResNet-50) backbone is leveraged within a deep feature extraction module to capture hierarchically rich and discriminative representations. To effectively integrate heterogeneous feature streams, we design a Multi-Branch Feature Fusion (MFF) module inspired by channel-wise attention mechanisms, which dynamically allocates contribution weights across branches through learned activation parameters. Extensive experiments on the cross-view multi-condition CASIA-B benchmark demonstrate that our method achieves a Rank-1 accuracy of 94.52\% under normal walking, with the best recognition performance among skeleton-based methods for the coat-wearing condition. On a challenging self-collected outdoor dataset, our approach surpasses existing state-of-the-art methods by 4.1\% in overall accuracy, exhibiting robust generalization from controlled indoor environments to unconstrained open-world scenarios.
\end{abstract}

\section{Introduction}

Biometric identification technologies have undergone rapid advancement in recent years, driven by growing demands in public safety, intelligent surveillance, and access control~\cite{sepas2023deep, mogan2023gait}. Among the diverse palette of biometric modalities---including fingerprint, iris, and face recognition---gait recognition occupies a distinctive niche owing to its non-invasive nature and capability to operate at considerable distances without requiring subject cooperation~\cite{yu2005hough, li2020gait}. As a behavioral biometric, gait encodes an individual's unique locomotion patterns, encompassing stride length, walking cadence, joint articulation, and overall body posture, which collectively form a rich signature that is inherently difficult to imitate or conceal~\cite{song2019gaitnet}.

The significance of gait recognition is further underscored by its applicability in scenarios where conventional biometrics falter. In far-field surveillance environments, face and iris features are often degraded by low resolution and unfavorable imaging conditions~\cite{kong2023deep}. Gait patterns, by contrast, remain discernible even at distances exceeding 50 meters, rendering gait recognition an indispensable complement to existing identification pipelines~\cite{zhu2022gaitref}. Furthermore, gait can be captured unobtrusively from standard surveillance footage without any specialized hardware or subject awareness, making it particularly attractive for forensic investigation and real-time threat detection~\cite{zheng2022gait}.

Traditional gait recognition methods predominantly rely on handcrafted feature descriptors informed by domain-specific prior knowledge. Yu~\emph{et~al.}~\cite{yu2005hough} employed Hough-transform-based descriptors to characterize leg movement by fitting joint angle trajectories into feature vectors. While pioneering, such approaches suffer from limited expressiveness and sensitivity to environmental perturbations, making them inadequate for the demands of modern surveillance systems that must operate robustly across diverse conditions~\cite{liao2017ptsn}.

The advent of deep learning has catalyzed a paradigm shift in gait recognition research. Contemporary methods have demonstrated remarkable proficiency in automatically learning hierarchical feature representations from raw gait data, substantially surpassing the performance ceiling of traditional approaches~\cite{he2016deep, sepas2023deep}. Current mainstream methods can be broadly categorized into appearance-based and model-based approaches.

\textbf{Appearance-based methods} typically operate on silhouette sequences extracted from walking videos. GaitSet~\cite{chao2022gaitset} innovatively treats cross-view gait sequences as unordered sets for temporal information extraction, achieving robust cross-view recognition. GaitPart~\cite{fan2020gaitpart} explores fine-grained local details within input silhouettes and employs temporal modeling to capture motion dynamics. GaitDAN~\cite{huang2024gaitdan} addresses domain shift arising from viewpoint variation through adversarial adaptation, mining rich spatiotemporal information for cross-view scenarios. Despite their strong overall performance, silhouette-based methods exhibit inherent limitations in anti-interference capability. They are susceptible to confusion between body contours and carried objects, and their recognition accuracy degrades significantly when subjects are partially occluded or overlapping~\cite{dou2022gaitgci}.

\textbf{Model-based methods} leverage skeletal keypoint representations to provide a more intrinsic characterization of human motion. Wang~\emph{et~al.}~\cite{wang2018mgn} proposed the Multi-Granularity Network (MGN) to exploit gait features at varying spatial scales. GaitGraph~\cite{teepe2021gaitgraph} pioneered the integration of Graph Convolutional Networks (GCNs) with skeletal poses, demonstrating the potential of topological structure modeling for gait recognition. GPGait~\cite{fu2023gpgait} introduced a generalized pose-based framework with masking operations to encourage fine-grained feature learning. More recently, SkeletonGait++~\cite{fan2024skeletongait} combines skeleton and silhouette streams in a dual-branch architecture that leverages complementary information sources~\cite{peng2024learning}.

An emerging trend in the field involves transitioning evaluation from controlled laboratory datasets to open-world scenarios. Fan~\emph{et~al.}~\cite{fan2023opengait} developed GaitBase, a structurally streamlined yet robust baseline within the OpenGait framework. BigGait~\cite{ye2024biggait} harnesses large vision models for unsupervised gait feature extraction, demonstrating strong performance on the outdoor GREW dataset. However, a substantial performance gap persists between indoor and outdoor settings; many methods experience accuracy drops exceeding 15\% when transitioning to unconstrained environments~\cite{zheng2024parsing, wang2023dygait}.

In the broader context of visual understanding and feature representation learning, recent advances in document intelligence and multi-modal learning have introduced powerful methodologies for structured feature extraction and cross-modal fusion that inspire our approach. Tang~\emph{et~al.}~\cite{tang2022fewcould} demonstrated that selective feature sampling and grouping can outperform exhaustive feature utilization for text detection, a principle analogous to our selective multi-branch fusion strategy. The attention-based fusion mechanisms explored in TextSquare~\cite{tang2024textsquare} and DocPedia~\cite{feng2023docpedia} for integrating heterogeneous visual and textual features share conceptual similarities with our channel-wise feature reweighting approach. Furthermore, the multi-modal interaction paradigms in UniDoc~\cite{feng2023unidoc} and MTVQA~\cite{tang2025mtvqa} for simultaneous detection, recognition, and understanding provide valuable insights into designing architectures that must jointly process and fuse diverse information streams---a challenge equally central to multi-feature gait recognition. The hierarchical feature extraction strategies in SPTS~v2~\cite{liu2023sptsv2} and optimal bounding box learning via reinforcement in~\cite{tang2022optimalboxes} further reinforce the importance of adaptive, learnable feature selection mechanisms that we adopt in our MFF module.

To address the limitations of existing gait recognition methods in complex real-world scenarios, we propose a skeleton-based multi-feature fusion framework that comprehensively extracts and integrates multiple discriminative characteristics of human locomotion. Our key contributions are:

\begin{itemize}
\item We design a complete gait recognition pipeline that employs HRNet~\cite{sun2019hrnet} for robust keypoint estimation under low-resolution conditions, followed by gait cycle extraction using leg movement similarity analysis, establishing a solid foundation for subsequent feature modeling.

\item We construct three complementary feature branches---body proportion, gait velocity, and skeletal motion---that capture static anthropometric characteristics, temporal dynamics, and kinematic properties respectively, providing a comprehensive representation of individual identity.

\item We propose a Multi-Branch Feature Fusion (MFF) module that performs mid-level feature integration through a channel attention-inspired mechanism with dynamic weight allocation, enabling effective exploitation of complementary advantages across branches.

\item Extensive experiments on the CASIA-B benchmark and a self-collected outdoor dataset demonstrate that our method achieves state-of-the-art performance among skeleton-based approaches, with particularly strong robustness under covariate interference and in open-world scenarios.
\end{itemize}

\section{Related Work}

\subsection{Appearance-Based Gait Recognition}

Appearance-based gait recognition methods predominantly operate on silhouette sequences, which encode the overall body shape and motion pattern of walking subjects. Silhouette representations dominate the field, constituting approximately 81\% of published methods~\cite{sepas2023deep}. A central research focus has been on effective spatiotemporal feature extraction and modeling from these sequences.

GaitSet~\cite{chao2022gaitset} introduced a set-based paradigm that treats gait frames as unordered sets, compressing frame-level spatial features into a compact representation invariant to temporal ordering. GaitPart~\cite{fan2020gaitpart} decomposed gait sequences into temporal parts and applied dedicated micro-motion capture modules to model local dynamics, revealing the importance of fine-grained temporal modeling. Building upon these foundations, GaitDAN~\cite{huang2024gaitdan} specifically targeted cross-view challenges by leveraging adversarial domain adaptation with neighbor-pair networks to mine discriminative spatiotemporal features under view changes.

More recent works have pushed the boundaries of silhouette-based recognition. GaitGCI~\cite{dou2022gaitgci} employed generative counterfactual intervention to disentangle identity-relevant features from confounding factors. Context-sensitive temporal feature learning~\cite{huang2021context} demonstrated the value of adaptive temporal modeling. DyGait~\cite{wang2023dygait} proposed dynamic representations that adaptively capture motion patterns, achieving strong results on multiple benchmarks. LidarGait~\cite{shen2023lidargait} extended gait recognition to 3D point cloud inputs, opening new avenues for modality-agnostic recognition.

Despite their effectiveness, silhouette-based methods remain vulnerable to appearance-altering covariates such as clothing changes and carried objects, as the outer contour can be significantly distorted by these factors~\cite{lin2021gait}. This fundamental limitation motivates the exploration of skeleton-based approaches.

\subsection{Skeleton-Based Gait Recognition}

Skeleton-based methods represent gait through body joint coordinates, offering an inherently more robust representation against appearance-level interference. Early skeleton-based approaches focused on manually designed features. Liao~\emph{et~al.}~\cite{liao2017ptsn} proposed the Pose-based Temporal-Spatial Network (PTSN) that explicitly models temporal and spatial relationships among joint positions, achieving competitive performance under carrying and clothing variations.

The integration of Graph Convolutional Networks (GCNs) has significantly advanced skeleton-based gait recognition. GaitGraph~\cite{teepe2021gaitgraph} combined skeleton poses with GCN architectures, establishing a modern model-based baseline. Subsequent work by Teepe~\emph{et~al.}~\cite{teepe2022towards} provided deeper analyses of the design choices critical for skeleton-based approaches. Symmetry-driven hyper feature GCNs have further exploited the bilateral symmetry of human locomotion for enhanced discrimination.

GPGait~\cite{fu2023gpgait} introduced a unified pose-based framework with masking operations to encourage the model to focus on fine-grained feature extraction. Fan~\emph{et~al.}~\cite{fan2024skeletongait} developed SkeletonGait++, which innovatively combines skeleton map representations with silhouette features in a dual-stream architecture, demonstrating the complementary value of multi-source gait information~\cite{peng2024learning}. BigGait~\cite{ye2024biggait} leveraged large-scale vision models for unsupervised gait representation learning, achieving state-of-the-art results on in-the-wild datasets.

However, existing skeleton-based methods often underutilize the rich biometric information available in skeletal data, typically focusing on either spatial or temporal features in isolation. Our work addresses this gap by simultaneously extracting body proportion, gait velocity, and skeletal motion features, and fusing them through a principled attention-based mechanism.

\subsection{Feature Fusion in Biometric Recognition}

Feature fusion strategies play a pivotal role in enhancing biometric recognition systems by combining complementary information sources. Wang~\emph{et~al.}~\cite{wang2018mgn} proposed multi-granularity feature fusion for person re-identification, concatenating features at different spatial scales. Guo~\emph{et~al.}~\cite{guo2022person} embedded attention mechanisms into global and local branches for person re-identification, demonstrating improved robustness under partial occlusion.

In the realm of multi-modal and multi-feature learning, significant insights have been drawn from adjacent fields. The concept of adaptive feature selection has been extensively explored in document understanding~\cite{tang2022fewcould, tang2024textsquare}, where selective feature grouping and attention-based weighting have proven more effective than exhaustive feature aggregation. Multi-modal fusion architectures in visual text understanding~\cite{zhao2024harmonizing, zhao2024tabpedia} provide compelling evidence that heterogeneous feature streams benefit from intermediate-level fusion with learned channel-wise attention, rather than naive late fusion through concatenation. The bounding box tokenization approach in~\cite{lu2025boundingbox} and the heterogeneous anchor prompting in Dolphin~\cite{feng2025dolphin} demonstrate how structured spatial information can be effectively integrated with semantic features---a paradigm that resonates with our fusion of spatial body proportion features with temporal gait dynamics.

The hierarchical feature fusion strategies employed in PARGO~\cite{wang2024pargo} and the multi-modal in-context learning framework~\cite{zhao2023mmicl} further motivate our design choice of fusing features at the network's intermediate layers rather than at the output level. MCTBench~\cite{shan2024mctbench} and WildDoc~\cite{wang2025wilddoc} also highlight the importance of comprehensive benchmarking across diverse conditions, paralleling our evaluation strategy that spans both controlled and open-world gait recognition scenarios.

Traditional fusion approaches in gait recognition typically perform late fusion through fully connected layers at the network output, neglecting inter-feature correlations and complementarity. Inspired by channel attention mechanisms~\cite{lei2020hierarchical}, our Multi-Branch Feature Fusion module performs intermediate-level fusion with dynamic weight allocation, enabling effective exploitation of feature complementarity while maintaining computational efficiency.

\section{Method}

\begin{figure*}[t]
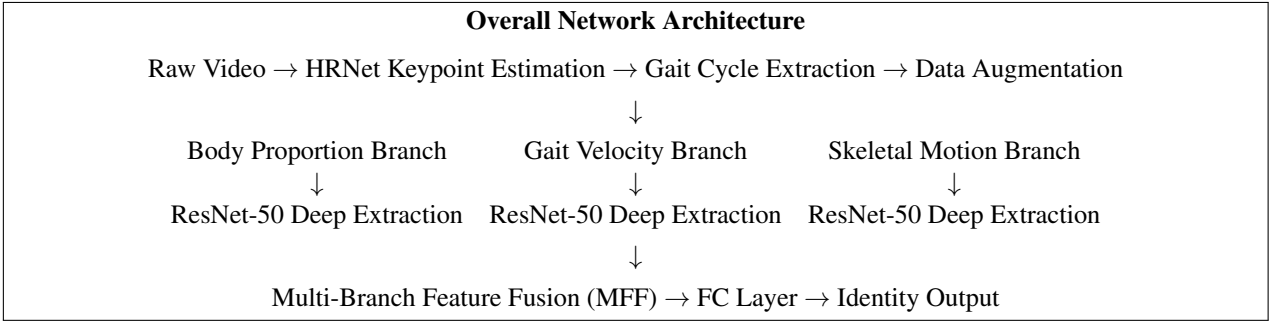

\centering
\fbox{\parbox{0.95\textwidth}{\centering \textbf{Overall Network Architecture}\\[6pt]
Raw Video $\rightarrow$ HRNet Keypoint Estimation $\rightarrow$ Gait Cycle Extraction $\rightarrow$ Data Augmentation\\[4pt]
$\downarrow$\\[4pt]
$\begin{array}{ccc}
\text{Body Proportion Branch} & \text{Gait Velocity Branch} & \text{Skeletal Motion Branch} \\
\downarrow & \downarrow & \downarrow \\
\text{ResNet-50 Deep Extraction} & \text{ResNet-50 Deep Extraction} & \text{ResNet-50 Deep Extraction}
\end{array}$\\[4pt]
$\downarrow$\\[4pt]
Multi-Branch Feature Fusion (MFF) $\rightarrow$ FC Layer $\rightarrow$ Identity Output}}
\caption{Overview of the proposed gait recognition framework. Raw walking videos are first processed by HRNet for keypoint detection, followed by gait cycle extraction. Three complementary feature branches are constructed and independently refined by ResNet-50 deep feature extraction modules. The Multi-Branch Feature Fusion (MFF) module integrates all branches through dynamic attention-based weighting to produce the final identity prediction.}
\label{fig:overview}
\end{figure*}

We present a comprehensive gait recognition framework that integrates skeleton-based multi-feature extraction with an attention-inspired fusion module. The overall pipeline, illustrated in Figure~\ref{fig:overview}, comprises four main stages: (1) skeletal keypoint estimation via HRNet, (2) gait cycle extraction, (3) multi-branch feature modeling, and (4) deep feature extraction and fusion through the MFF module.

\subsection{Skeletal Keypoint Estimation via HRNet}

Accurate skeletal keypoint estimation serves as the foundation of our entire recognition pipeline. We adopt the High-Resolution Network (HRNet)~\cite{sun2019hrnet} for human pose estimation, following the annotation protocol of the MPII Human Pose dataset with 16 keypoints spanning the head, upper body, and lower extremities.

Our approach follows a top-down paradigm comprising two stages: target person detection followed by single-person keypoint estimation. This methodology maximizes the utilization of holistic spatial information as prior knowledge, yielding superior accuracy compared to bottom-up alternatives~\cite{cao2017realtime}. The top-down approach also exhibits stronger robustness to occlusion, making it well-suited for multi-person complex scenarios.

A critical challenge in gait recognition is that surveillance cameras typically capture subjects at considerable distances, resulting in low-resolution inputs. Conventional pose estimation architectures tend to progressively downsample feature maps through successive convolutional stages, causing irreversible loss of high-resolution spatial details. HRNet addresses this limitation by maintaining parallel multi-resolution subnetworks throughout the network computation. Rather than applying standard convolutions that reduce spatial dimensions, HRNet employs upsampling and transposed convolutions to preserve consistent width and height information, with the final output produced from the highest-resolution branch.

The multi-resolution fusion mechanism enables deep information extraction across scales, significantly enhancing the precision of keypoint localization. This capability is particularly crucial for our application, where the network must maintain reliable performance even with $256 \times 256$ pixel inputs that reflect the low-resolution characteristics of typical surveillance footage.

\subsection{Gait Cycle Extraction}

Human gait analysis operates on the fundamental temporal unit of the gait cycle, defined as the interval between two consecutive contacts of the same foot with the ground. We extract gait cycles by computing the frame-to-frame similarity of lower-limb skeletal configurations.

Specifically, we employ the Hamming distance metric to estimate the similarity of lower-limb keypoint positions across consecutive frames. Using the first frame of the target person's gait sequence as a reference, we compute the Hamming distance between the reference and subsequent frames, generating a similarity waveform. The periodic nature of walking produces a characteristic oscillatory pattern in this similarity signal, where adjacent troughs correspond to half-gait-cycle intervals.

For a representative walking sequence, the similarity analysis reveals that frames exhibiting maximum similarity to the first frame (e.g., frames 12, 24, and 36) are separated by intervals of 12 frames, indicating a half-gait-cycle duration of 12 frames and a full cycle of 24 frames. This temporal structure serves as the basis for selecting the temporal window for subsequent feature extraction. To reduce computational overhead, we compute the sample mean of gait cycle durations across training batches and adopt twice the full gait cycle as the standard temporal stride. When new subjects are introduced, this hyperparameter is recalibrated to maintain optimal temporal coverage.

\subsection{Multi-Branch Feature Modeling}

From the HRNet-extracted skeletal keypoints, we construct a spatiotemporal graph sequence representing the human body skeleton:
\begin{equation}
X = \{x \in \mathbb{R}^{C \times T \times I}\},
\end{equation}
where $C$ denotes the coordinate dimensionality of joint positions, $T$ represents the temporal frame index, and $I$ indexes the joint keypoints. This representation serves as input to three complementary feature branches.

\subsubsection{Body Proportion Branch}

Body shape characteristics exhibit greater temporal stability compared to dynamic gait features, compensating for the sensitivity of motion features to acquisition conditions. To comprehensively encode relative positional information among keypoints, we introduce adjacency features and body proportion characteristics:
\begin{equation}
\text{Rate} = \left\{h_i \,|\, i = 1, 2, \ldots, I\right\},
\end{equation}
where $h_i = c[:,\!:\!,i]$ denotes the relative position of keypoint $x[:,\!:\!,i]$ with respect to the skeleton center $x[:,\!:\!,c]$, defined as the midpoint of the line connecting the thorax and pelvis. The skeletal adjacency features are concatenated with the original keypoint position features via matrix concatenation, forming the proportion branch input. This branch captures the subject's body proportion characteristics, including overall build dimensions, relative joint positions during movement, and range of motion.

\subsubsection{Gait Velocity Branch}

Temporal dynamics carry significant discriminative information for gait recognition. We design a velocity branch that combines short-sequence and per-frame differential features to capture multi-scale temporal dynamics:
\begin{equation}
F = \left\{f_t \,|\, t = 1, 2, \ldots, T\right\},
\end{equation}
\begin{equation}
E = \left\{e_t \,|\, t = 1, 2, \ldots, T\right\},
\end{equation}
where $f_t = x[:,t+6,:] - x[:,t,:]$ represents the short-period velocity capturing displacement over 6-frame intervals, and $e_t = x[:,t+1,:] - x[:,t,:]$ represents the instantaneous velocity from adjacent frame differences. The fusion of $F$ and $E$ constitutes the velocity branch, encoding both instantaneous and short-period temporal dynamics that characterize the subject's unique locomotion rhythm.

\subsubsection{Skeletal Motion Branch}

To provide an intrinsic kinematic characterization of walking, we construct a skeletal motion branch incorporating bone lengths and articulation angles:
\begin{equation}
L = \left\{l_i \,|\, i = 1, 2, \ldots, I\right\},
\end{equation}
\begin{equation}
A = \left\{a_i \,|\, i = 1, 2, \ldots, I\right\},
\end{equation}
where $l_i = x[:,\!:\!,i] - x[:,\!:\!,\text{adj}(i)]$ represents the length of the bone connecting keypoint $i$ to its adjacent keypoint $\text{adj}(i)$, and
\begin{equation}
a_i = \arccos\left(\frac{l_{i,y}}{\sqrt{l_{i,x}^2 + l_{i,y}^2}}\right)
\end{equation}
computes the angle between bone $i$ and the horizontal direction. This branch provides a direct and intuitive description of the subject's gait posture, stride length, and cadence through the skeletal representation.

Through the above three-branch decomposition, the initial keypoint information is expanded into semantically enriched and complementary feature streams, each capturing distinct aspects of the subject's identity.

\subsection{Deep Feature Extraction via ResNet-50}

Each feature branch is independently processed through a deep feature extraction module built upon the 50-layer Residual Network (ResNet-50)~\cite{he2016deep}. The residual architecture introduces a novel layer-wise information transfer mechanism through skip connections:
\begin{equation}
F(x) = W_2 \sigma(W_1 x),
\end{equation}
where $W_1$ and $W_2$ denote the weight matrices of successive convolutional layers, and $\sigma$ represents the Rectified Linear Unit (ReLU) activation function. The core insight is that residual learning reformulates the optimization target as $F(x) = H(x) - x$, where training drives $F(x) \to 0$ to achieve identity mappings, enabling effective training of substantially deeper networks without gradient degradation.

Our deep feature extraction module commences with a $7 \times 7$ convolutional layer for preliminary global feature extraction, which simultaneously reduces network parameters. This is followed by sequences of optimized residual blocks, each containing a $3 \times 3$ convolutional layer flanked by two $1 \times 1$ convolutional layers for dimensionality alignment. This bottleneck architecture efficiently balances feature extraction depth with computational overhead, producing highly discriminative feature representations for each branch.

The choice of ResNet-50 represents a deliberate trade-off between model capacity and practical constraints. While deeper variants (e.g., ResNet-101, ResNet-152) offer marginally greater representational power, ResNet-50 provides sufficient feature extraction capability with significantly fewer parameters and faster training convergence, making it well-suited for the skeletal feature dimensions in our application.

\subsection{Multi-Branch Feature Fusion Module}

The Multi-Branch Feature Fusion (MFF) module is the centerpiece of our architecture, designed to integrate heterogeneous feature streams through a channel attention-inspired mechanism. Unlike conventional late-fusion approaches that simply concatenate features at the network output, MFF performs intermediate-level fusion with dynamic weight allocation, enabling principled exploitation of inter-branch complementarity.

The fusion process consists of five sequential steps:

\textbf{Step 1: Spatial Information Aggregation.}
The proportion branch and skeletal motion branch are concatenated at the input level via matrix concatenation to form a unified spatial feature representation $f_w$. The velocity branch produces feature maps $f_v$. Due to dimensionality differences between $f_w$ and $f_v$, global average pooling is applied independently to each, and the results are concatenated to yield a joint feature $f_c$.

\textbf{Step 2: Dimensionality Reduction.}
The concatenated feature $f_c$ is projected into a lower-dimensional space to reduce subsequent computational cost and improve efficiency:
\begin{equation}
f_a = W \cdot f_c + b,
\end{equation}
where $f_a$ denotes the fused feature, $W$ is the learnable weight matrix, and $b$ is the bias term.

\textbf{Step 3: Branch-wise Excitation.}
An activation function maps the fused feature $f_a$ back to each branch's feature space, computing excitation values that quantify each branch's relevance:
\begin{equation}
e_w = \sigma(W \cdot f_a + b),
\end{equation}
where $e_w$ represents the excitation value for the spatial information branch and $\sigma$ denotes the sigmoid activation function.

\textbf{Step 4: Attention-Weighted Recalibration.}
The excitation values serve as attention weights for their respective branches. They are expanded to match the dimensionality of $f_w$ and $f_v$, and element-wise Hadamard products are computed to recalibrate each branch's feature maps according to their task relevance.

\textbf{Step 5: Global Feature Aggregation.}
Average pooling is applied to obtain spatial and velocity features, which are concatenated to form the final global feature representation. This global average pooling compresses feature maps from $H \times W \times C$ to $1 \times 1 \times C$, preserving channel information while reducing spatial redundancy, which simultaneously prevents overfitting and stabilizes training.

The MFF module can be interpreted as a channel-wise attention mechanism that pre-assigns importance weights to different branches, selectively amplifying features beneficial to the current recognition task while suppressing less informative channels. The resulting global feature encompasses filtered, weighted contributions from all branches, preserving their individual strengths while minimizing redundant or conflicting information.

\subsection{Analysis of Feature Complementarity}

The design rationale for our three-branch architecture stems from a careful analysis of feature complementarity:

\begin{itemize}
\item \textbf{Body proportion features} are relatively stable and discriminative but lack dynamic walking characteristics. Their recognition accuracy is sensitive to the precision of the upstream keypoint estimation network.

\item \textbf{Gait features} (velocity and skeletal motion) encode temporal displacement and kinematic information that provides explicit, efficient, and diverse representations of locomotion. However, their discriminability may be compromised at extreme viewing angles (e.g., 0\degree~or 180\degree).

\item \textbf{Feature fusion} enables the body proportion branch to maintain stability under viewpoint changes or limited training samples, while the gait branches contribute more explicit and multi-faceted motion representations. The combined model thus achieves enhanced recognition performance and generalization across diverse operating conditions.
\end{itemize}

\section{Experiments}

We conduct comprehensive experiments to evaluate the proposed method across multiple dimensions: keypoint detection accuracy, individual branch contribution, multi-view robustness, comparison with state-of-the-art methods, model efficiency, and open-world generalization.

\subsection{Datasets}

\textbf{MPII Human Pose Dataset.}
The MPII dataset~\cite{kong2023deep} comprises approximately 25,000 annotated images covering more than 40,000 subjects across 410 activities, and serves as the benchmark for evaluating our HRNet-based keypoint detection module. It includes annotations for body part occlusion, 3D torso, and head orientation.

\textbf{CASIA-B Dataset.}
The CASIA-B dataset~\cite{yu2006framework} is the largest publicly available cross-view gait database, containing 13,640 video sequences from 124 subjects. Each subject is captured from 11 viewing angles (0\degree~to 180\degree~at 18\degree~intervals) under three walking conditions: normal walking (NM, 6 sequences), wearing a coat (CL, 2 sequences), and carrying a bag (BG, 2 sequences). Following the standard large-training-set protocol, we use 74 subjects for training and 50 for testing.

\textbf{Self-Collected Outdoor Dataset.}
To evaluate open-world generalization, we collected a dataset of 68 subjects captured by cameras mounted at 2.3~m height under uncontrolled outdoor conditions with natural lighting variation and dynamic backgrounds. Data collection covers 7 viewing angles (0\degree, 36\degree, 72\degree, 90\degree, 108\degree, 144\degree, 180\degree) and the same three walking conditions as CASIA-B (NM: 6 sequences, CL: 2 sequences, BG: 2 sequences).

\subsection{Implementation Details}

All experiments are conducted using the TensorFlow deep learning framework with Python 3.7 on an Intel Core i7-8550U CPU with 16~GB RAM, using CUDA 10.1. The HRNet backbone accepts $256 \times 256$ pixel input images. For the multi-branch fusion network, we set: batch size = 64, learning rate = 0.0001, learning rate decay = 0.01, dropout = 0.35, and train for 150 epochs. The loss function converges and stabilizes around epoch 130, and we select the model weights from that checkpoint.

\subsection{Keypoint Detection Evaluation}

We first validate the HRNet keypoint detection module on the MPII test set (6,619 images). The evaluation metric is the Head-normalized Probability of Correct Keypoint (PCKh):
\begin{equation}
f_{\text{PCKh}} = \frac{\sum_m \delta\left(\frac{d_m^i}{d_m^h} \leq T_k\right)}{\sum_m 1},
\end{equation}
where $d_m^i$ is the Euclidean distance between the predicted and ground-truth positions of the $i$-th keypoint for person $m$, $d_m^h$ is the head scale factor, $T_k = 0.01$ is the threshold, and $\delta$ is the indicator function.

At the $256 \times 256$ input resolution with $T_k = 0.01$, the average PCKh across all keypoints exceeds 83\%. Critically, the keypoint localization for lower extremities---most relevant for gait analysis---is particularly strong, with PCKh values exceeding 95\% for both ankle and knee joints. These results confirm that HRNet provides sufficiently accurate skeletal estimates for downstream gait recognition.

\subsection{Ablation Study}

We conduct ablation experiments on CASIA-B to quantify the contribution of each feature branch. Using the first 4 NM sequences per subject for enrollment and the remaining 6 sequences (2 NM, 2 BG, 2 CL) for verification, we systematically evaluate combinations of the three branches.

\begin{table}[t]
\centering
\caption{Ablation study on CASIA-B. Each row indicates the active feature branches. Bold values denote the best results.}
\label{tab:ablation}
\begin{tabular}{l|ccc|c}
\toprule
\textbf{Branch Combination} & \textbf{NM} & \textbf{BG} & \textbf{CL} & \textbf{Overall} \\
\midrule
Velocity only & 71.46 & 68.35 & 56.84 & 65.55 \\
Proportion + Skeleton & 71.89 & 70.98 & 62.46 & 68.44 \\
Proportion + Velocity & 86.31 & 76.83 & 68.46 & 77.20 \\
Skeleton + Velocity & 90.37 & 80.38 & 79.19 & 83.31 \\
All Three (Ours) & \textbf{94.52} & \textbf{85.92} & \textbf{82.97} & \textbf{87.81} \\
\bottomrule
\end{tabular}
\end{table}

As shown in Table~\ref{tab:ablation}, single-feature and dual-feature configurations yield substantially lower accuracy compared to the full three-branch fusion. The velocity-only baseline achieves merely 65.55\% overall, with particularly poor performance under the CL condition (56.84\%) due to reduced keypoint estimation accuracy when thick clothing is present. Incorporating the proportion branch with the velocity branch improves performance to 77.20\%, and adding the skeletal motion branch further boosts accuracy to 83.31\%. The complete three-branch fusion achieves the highest overall accuracy of 87.81\%, with the NM accuracy reaching 94.52\% and an overall improvement of 4.50\% over the next-best combination, conclusively demonstrating the complementary value of multi-feature integration.

\subsection{Cross-View Recognition}

\begin{table}[t]
\centering
\caption{Cross-view recognition rates (\%) on CASIA-B across 11 viewing angles. Bold values denote the best results per condition.}
\label{tab:crossview}
\resizebox{\columnwidth}{!}{
\begin{tabular}{c|ccc}
\toprule
\textbf{Angle} & \textbf{NM} & \textbf{BG} & \textbf{CL} \\
\midrule
0\degree & 94.33 & 84.29 & 81.14 \\
18\degree & 95.67 & 86.78 & 81.71 \\
36\degree & \textbf{97.36} & 86.78 & 84.24 \\
54\degree & 96.33 & 85.00 & 83.94 \\
72\degree & 94.67 & 84.87 & 82.69 \\
90\degree & 93.33 & 83.68 & 83.69 \\
108\degree & 94.12 & 84.79 & 84.64 \\
126\degree & 93.33 & 83.69 & \textbf{88.29} \\
144\degree & 92.67 & \textbf{87.79} & 83.04 \\
162\degree & 89.87 & 84.29 & 82.71 \\
180\degree & 89.14 & 87.79 & 81.14 \\
\bottomrule
\end{tabular}
}
\end{table}

Table~\ref{tab:crossview} presents the cross-view recognition results. The model demonstrates robust performance across all viewing angles, with peak NM accuracy of 97.36\% at 36\degree~and consistently high performance at nearby intervals. Even at the challenging 90\degree~angle, where partial body occlusion occurs, the NM recognition rate remains at 93.33\%, evidencing strong robustness to information loss from self-occlusion.

The model's performance at 0\degree~and 180\degree~is relatively lower but still exceeds 80\% across all conditions, as the reduced lateral movement visibility at frontal and rear views naturally diminishes gait discriminability. The BG group shows slight accuracy reduction due to misidentification of keypoints around the back and shoulders, while the CL group exhibits the most pronounced degradation at side views due to the overall body contour expansion caused by thick clothing. Nevertheless, the velocity branch's lower sensitivity to keypoint localization precision provides compensatory stability under these challenging conditions.

\subsection{Comparison with State-of-the-Art Methods}

\begin{table}[t]
\centering
\caption{Comparison with state-of-the-art methods on CASIA-B under the large-training-set setting. Bold values indicate best results.}
\label{tab:comparison}
\begin{tabular}{l|ccc|c}
\toprule
\textbf{Method} & \textbf{NM} & \textbf{BG} & \textbf{CL} & \textbf{Overall} \\
\midrule
GaitNet~\cite{lei2020hierarchical} & 91.5 & 85.7 & 58.9 & 78.7 \\
PTSN~\cite{liao2017ptsn} & 97.0 & 85.8 & 68.1 & 83.6 \\
GaitBase~\cite{fan2023opengait} & 96.9 & 72.8 & 86.4 & 89.2 \\
GaitSet~\cite{chao2022gaitset} & 95.0 & 87.2 & 70.4 & 84.2 \\
GPGait~\cite{fu2023gpgait} & 92.7 & 84.8 & 79.3 & 85.6 \\
BigGait~\cite{ye2024biggait} & 95.9 & 86.4 & 79.1 & 87.1 \\
SkeletonGait++~\cite{fan2024skeletongait} & 93.2 & 87.1 & 80.6 & 86.9 \\
\textbf{Ours} & \textbf{94.5} & \textbf{85.9} & \textbf{83.0} & \textbf{89.4} \\
\bottomrule
\end{tabular}
\end{table}

Table~\ref{tab:comparison} presents a comprehensive comparison with state-of-the-art methods spanning both skeleton-based (PTSN, GPGait, SkeletonGait++) and silhouette-based (GaitSet, BigGait, GaitBase) paradigms.

Our method achieves the highest overall accuracy of 89.4\%, surpassing all compared approaches. Under the CL condition, our approach demonstrates a decisive advantage with 83.0\% accuracy, outperforming PTSN by 14.9\%, GPGait by 3.7\%, and BigGait by 3.9\%. This substantial improvement under clothing variation---the most challenging covariate---validates our multi-feature fusion strategy, which effectively exploits the stability of body proportion features and the kinematic robustness of skeletal motion features to compensate for the degraded keypoint accuracy under thick clothing.

While PTSN achieves slightly higher NM accuracy (97.0\% vs. 94.5\%), its performance collapses dramatically under the CL condition (68.1\%), revealing an over-reliance on appearance-sensitive features. BigGait demonstrates competitive overall performance but suffers from weak suppression of low-frequency color noise in its unsupervised paradigm, reducing robustness to clothing changes.

\subsection{Model Efficiency Analysis}

\begin{table}[t]
\centering
\caption{Comparison of model parameters and overall accuracy on CASIA-B. Bold values indicate best results.}
\label{tab:params}
\begin{tabular}{l|c|c|c}
\toprule
\textbf{Method} & \textbf{Venue} & \textbf{Params (M)} & \textbf{Overall (\%)} \\
\midrule
GaitSet~\cite{chao2022gaitset} & TPAMI'22 & 4.28 & 84.2 \\
GPGait~\cite{fu2023gpgait} & ICCV'23 & 3.86 & 85.6 \\
GaitBase~\cite{fan2023opengait} & CVPR'23 & 11.76 & 86.4 \\
BigGait~\cite{ye2024biggait} & CVPR'24 & 30.82 & 87.1 \\
SkeletonGait++~\cite{fan2024skeletongait} & AAAI'24 & 8.49 & 86.9 \\
\textbf{Ours} & --- & \textbf{3.72} & \textbf{89.4} \\
\bottomrule
\end{tabular}
\end{table}

Table~\ref{tab:params} compares model efficiency in terms of parameter count. Our method achieves the \emph{smallest} model size (3.72M parameters) while attaining the \emph{highest} overall accuracy (89.4\%). This favorable efficiency-accuracy trade-off is attributable to several design choices: the bottleneck residual architecture constrains parameter growth within the deep feature extraction modules; the MFF module's dimensionality reduction mapping and global average pooling substantially compress feature redundancy; and the early fusion strategy avoids the parameter overhead of maintaining separate deep processing streams for late concatenation.

From a real-time processing perspective, the HRNet front-end achieves a single-frame pose estimation speed of 0.18~s, enabling near-real-time operation suitable for surveillance applications.

\subsection{Open-World Generalization}

\begin{table}[t]
\centering
\caption{Comparison on the self-collected outdoor dataset. Bold and underlined values denote the best and second-best results, respectively.}
\label{tab:outdoor}
\begin{tabular}{l|ccc|c}
\toprule
\textbf{Method} & \textbf{NM} & \textbf{BG} & \textbf{CL} & \textbf{Overall} \\
\midrule
GaitNet~\cite{lei2020hierarchical} & 75.7 & 63.2 & 50.7 & 65.2 \\
PTSN~\cite{liao2017ptsn} & 75.2 & 65.1 & 52.9 & 64.4 \\
GaitBase~\cite{fan2023opengait} & 87.1 & 82.2 & 69.8 & 79.7 \\
GaitSet~\cite{chao2022gaitset} & 70.3 & 69.8 & 58.5 & 66.2 \\
GPGait~\cite{fu2023gpgait} & 86.2 & 79.9 & 75.7 & 80.6 \\
BigGait~\cite{ye2024biggait} & 84.2 & 75.4 & 78.3 & 79.3 \\
SkeletonGait++~\cite{fan2024skeletongait} & \underline{87.5} & \underline{81.6} & 74.0 & \underline{81.0} \\
\textbf{Ours} & \textbf{87.4} & \textbf{83.9} & \textbf{84.2} & \textbf{85.1} \\
\bottomrule
\end{tabular}
\end{table}

The transition from controlled indoor to uncontrolled outdoor environments constitutes the most demanding evaluation scenario for gait recognition. As shown in Table~\ref{tab:outdoor}, most existing methods suffer accuracy drops exceeding 15\% in this transition, primarily due to real-world noise including complex occlusions, background clutter, and illumination variation.

Our method achieves the highest overall accuracy of 85.1\%, surpassing the second-best method (SkeletonGait++) by 4.1\%. Under the NM condition, our approach achieves 87.4\%, marginally behind SkeletonGait++ (87.5\%) by only 0.1\%. However, under the more challenging BG and CL conditions, our method demonstrates decisive advantages with 83.9\% and 84.2\%, respectively, outperforming the second-best approaches by 1.7\% and 5.9\%. The substantial improvement under the CL condition is particularly noteworthy, as clothing variation represents the dominant source of performance degradation in outdoor scenarios.

\begin{table}[t]
\centering
\caption{Multi-angle recognition rates (\%) on the self-collected outdoor dataset. Bold values denote the best results per condition.}
\label{tab:outdoor_angle}
\begin{tabular}{c|ccc}
\toprule
\textbf{Angle} & \textbf{NM} & \textbf{BG} & \textbf{CL} \\
\midrule
0\degree & 87.04 & 83.33 & 81.48 \\
36\degree & 86.38 & 85.19 & \textbf{92.59} \\
72\degree & \textbf{87.04} & \textbf{87.04} & 88.89 \\
90\degree & 86.38 & 81.48 & 79.63 \\
108\degree & 88.89 & 85.19 & 87.04 \\
144\degree & 86.38 & 83.33 & 87.04 \\
180\degree & 83.33 & 80.63 & 83.33 \\
\bottomrule
\end{tabular}
\end{table}

Table~\ref{tab:outdoor_angle} further details the multi-angle performance on the outdoor dataset. The recognition rates peak around 36\degree~and exhibit a gradual decline at extreme angles, consistent with the CASIA-B results. Despite the additional challenges posed by the overhead camera perspective and reduced lower-limb visibility, the network maintains over 80\% accuracy across all angles and conditions, confirming strong anti-interference capability under real-world operating conditions.

These results collectively demonstrate that our multi-branch feature fusion approach effectively bridges the indoor-outdoor performance gap that plagues many existing methods, establishing its practical viability for real-world surveillance deployment.

\section{Conclusion}

We have presented a skeleton-based multi-feature fusion framework for robust gait recognition that effectively addresses the challenges of covariate interference and environmental variability in practical surveillance applications. Our approach employs HRNet for accurate keypoint estimation under low-resolution conditions, extracts complementary body proportion, gait velocity, and skeletal motion features through a three-branch architecture, and integrates them via the proposed Multi-Branch Feature Fusion (MFF) module with dynamic attention-based weight allocation. Extensive experiments on the CASIA-B benchmark demonstrate state-of-the-art performance among skeleton-based methods, with a normal-walking accuracy of 94.52\% and particularly strong resilience to clothing variation (83.0\%), achieved with the smallest model footprint of 3.72M parameters. On a challenging self-collected outdoor dataset, our method surpasses existing approaches by 4.1\% in overall accuracy, demonstrating robust generalization from controlled to unconstrained environments and establishing practical viability for real-world deployment in intelligent surveillance and biometric identification systems.


\clearpage

{\small
\bibliographystyle{ieee_fullname}
\bibliography{references}
}

\end{document}